\documentclass[11pt]{article}

\usepackage[]{acl}

\usepackage{times}
\usepackage{latexsym}
\usepackage[T1]{fontenc}
\usepackage[utf8]{inputenc}
\usepackage{microtype}
\usepackage{inconsolata}
\usepackage{graphicx}
\usepackage{booktabs}
\usepackage{tabularx}
\usepackage{multirow}
\usepackage{amsmath}
\usepackage{makecell}

\newcolumntype{Y}{>{\raggedright\arraybackslash}X}

\newcommand{\dais}{\textsc{DAIS}}

\title{DAIS: Dependency-Aware Intermediate QA Supervision for Complex Reasoning}

\author {
    Yu Wang\textsuperscript{†1},
    Ming Fan\textsuperscript{*1},
    Xicheng Zhang\textsuperscript{1},
    Zhiyong Li\textsuperscript{1},\\
    \textbf{Zhihu Wang}\textsuperscript{2},
    \textbf{Caiyue Xu}\textsuperscript{2},
    \textbf{Dahai Hu}\textsuperscript{2},
    \textbf{Ting Liu}\textsuperscript{1}\\
    \textsuperscript{1}Xi’an Jiaotong University\quad
    \textsuperscript{2}Huawei Technologies Ltd.
}

\begin{document}
\maketitle
\renewcommand{\thefootnote}{}
\footnotetext{\textsuperscript{†}Work done during an internship at Huawei Technologies Ltd.}
\footnotetext{\textsuperscript{*}Corresponding Author}
\renewcommand{\thefootnote}{\arabic{footnote}}
\begin{abstract}
Chain-of-thought (CoT) supervision exposes intermediate rationales, but flat rationale targets usually optimize a single reasoning sequence and provide limited supervision on how local conclusions should support later decisions.
We introduce Dependency-Aware Intermediate QA Supervision (\dais{}), a training-time framework that converts filtered teacher rationales into stage-level QA records.
Each intermediate record predicts a local answer conditioned on the previous states needed for that decision, while the final-answer record keeps the original task format; evaluation therefore uses only the original input and optional context.
Across GDPR, AIACT, MedQA, and FOLIO with multiple Qwen backbones, DAIS improves average final-answer accuracy over answer-only, flat chain-of-thought, and independent-QA baselines.
On policy-compliance benchmarks, it achieves a largest gain of 5.6\% and an average gain of 4.2\% over the strongest non-DAIS baseline.
Controlled ablations show that valid previous-state conditioning contributes beyond longer targets or additional intermediate text, supporting dependency-conditioned intermediate QA as a lightweight auxiliary supervision signal for standard final-answer inference.
\end{abstract}

\section{Introduction}
\begin{figure*}[t]
    \centering
    \includegraphics[width=0.95\linewidth]{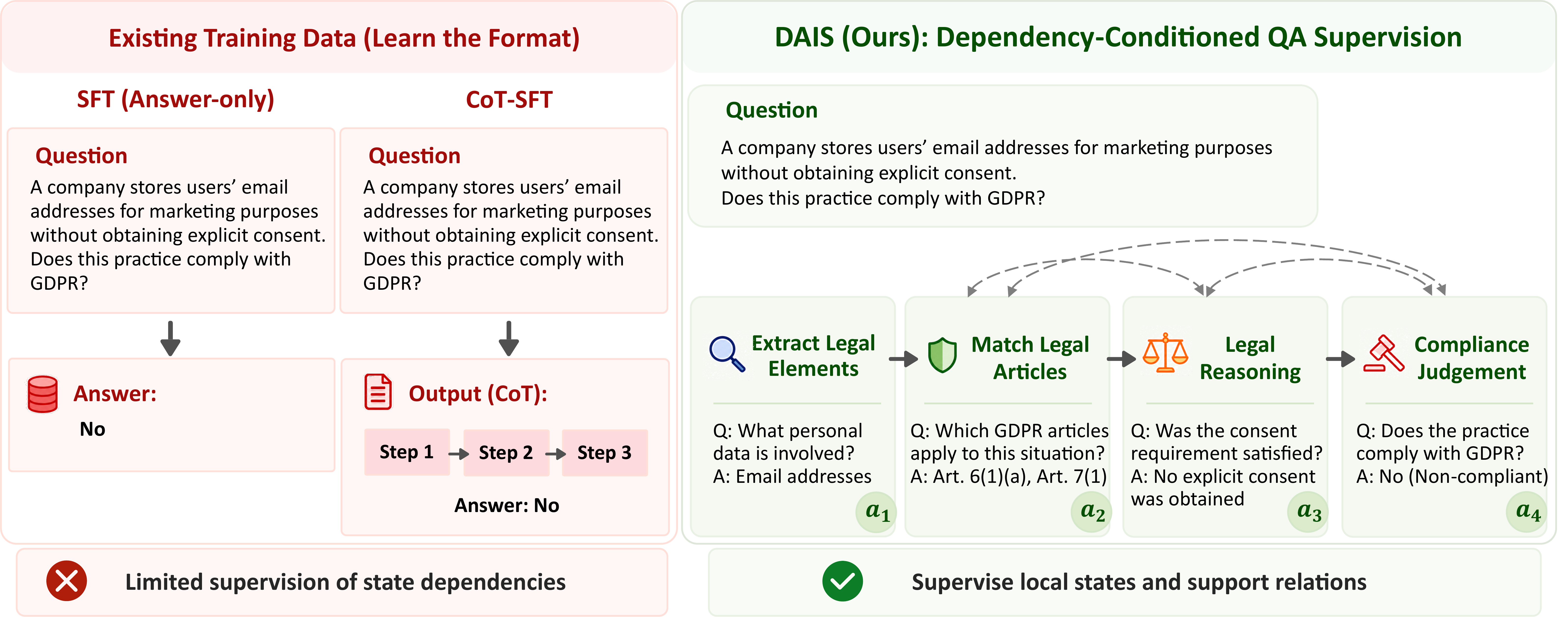}
    \caption{
Comparison between conventional supervised targets and \dais{}.
Answer-only SFT provides only the final label, flat CoT-SFT imitates a monolithic rationale, and \dais{} restructures the rationale into dependency-conditioned intermediate QA records used as auxiliary training supervision.
The final-answer record remains in the original task format, so evaluation can use standard direct final-answer inference.
}
    \label{fig:intro_compare}
\end{figure*}

Large language models (LLMs) have demonstrated strong performance on a wide range of reasoning tasks.
A key factor behind these advances is the use of intermediate reasoning before producing the final answer.
Chain-of-thought (CoT) prompting and supervision encourage models to generate explicit rationales and have become standard techniques for eliciting such reasoning behavior~\citep{wei2022chain,kojima2022large}.
Subsequent work further improves this paradigm through self-consistency, task decomposition, and tool-augmented reasoning~\citep{wang2023selfconsistency,zhou2023least,press2023measuring,yao2023react}.

Despite these gains, flat rationale supervision offers limited control over what is learned.
As illustrated by Figure~\ref{fig:intro_compare}, answer-only SFT supervises only the final label, while CoT-SFT adds intermediate text but optimizes it as one sequence.
This format does not represent local objectives or previous-state support relations as conditioning variables, so CoT gains can be entangled with target length, style, step markers, or answer-template imitation.

Intermediate-supervision methods make reasoning steps more explicit through rationale supervision~\citep{wei2022chain,kojima2022large}, decomposition~\citep{zhou2023least,press2023measuring}, and process supervision~\citep{uesato2022solving,lightman2023lets}, sometimes with inference-time search or verification.
Here, we focus on standard SFT, where intermediate targets often remain flat rationales, ordered steps, or local records without previous-state conditioning.
This motivates reorganizing teacher rationales into local QA records with previous-state support context while preserving direct final-answer inference. The key challenge is to use such dependencies as a training signal without turning them into an inference-time requirement.

To this end, we propose Dependency-Aware Intermediate QA Supervision (\dais{}).
\dais{} converts filtered teacher-generated CoT rationales into stage-level question-answer records.
Each intermediate record targets a local subtask extracted from the rationale, and later records are conditioned on selected earlier subtask answers that serve as support context for the current decision.
The original final-answer record is kept separately and remains in the standard task format.
Therefore, \dais{} can be trained with ordinary supervised fine-tuning and evaluated with direct final-answer prompting.

This design is complementary to inference-time decomposition methods.
\dais{} does not require a controller, search procedure, external verifier, gold intermediate state, or architecture change at evaluation time.
Instead, it asks whether dependency-conditioned intermediate targets provide a better supervised fine-tuning signal for standard final-answer prediction.

We evaluate \dais{} across four reasoning benchmarks spanning policy compliance, medical question answering, and logical reasoning, using four representative backbone models.
Experiments show that \dais{} consistently improves final-answer accuracy.
Notably, \dais{} achieves a largest gain of 5.6\% over the strongest non-\dais{} baseline on AIACT, with an average gain of 4.2\% on policy-oriented benchmarks.
These results demonstrate that bridging intermediate supervision and dependency-aware reasoning can yield substantial gains, especially in domains where final decisions depend on correctly using earlier reasoning states to evaluate applicable rules and constraints.

Our contributions are:
\begin{itemize}
    \item We propose \dais{}, a training-time dependency-conditioned intermediate supervision framework that converts teacher CoT rationales into stage-level QA records.
    \item We provide a lightweight SFT data-construction pipeline that preserves the standard direct final-answer inference setting.
    \item We show across four benchmarks and multiple Qwen backbones that valid previous-state conditioning improves final-answer accuracy beyond answer-only, flat CoT, independent-QA, order-only, length-matched, and corrupted-state controls.
\end{itemize}

\section{Related Work}
\label{sec:related_work}

\begin{figure*}[t]
    \centering
    \includegraphics[width=\linewidth]{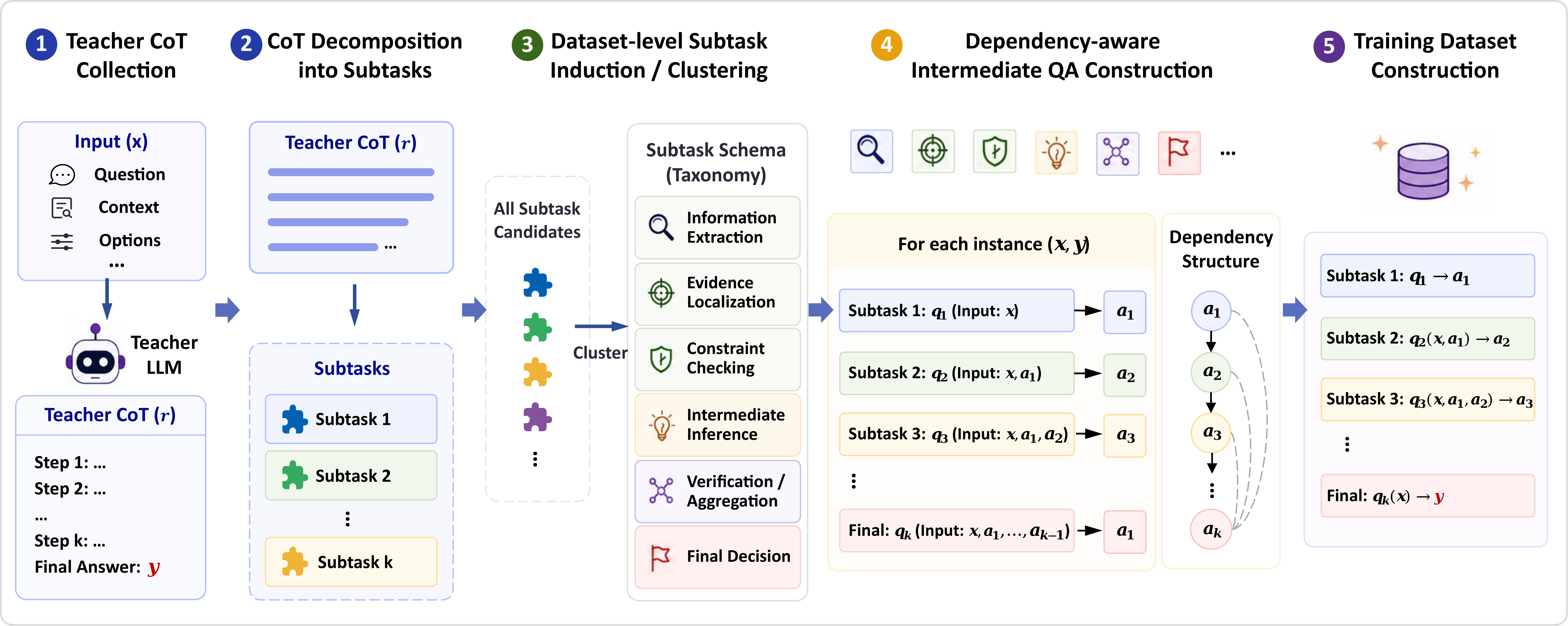}
\caption{
Overview of \dais{} data construction.
Teacher CoT rationales are decomposed into subtasks, normalized into dataset-level schemas, and converted into stage-level QA records.
Later intermediate records receive serialized previous subtask outputs as context, whereas the final-answer record remains the original final-task record and does not receive constructed intermediate states.
}
    \label{fig:methods}
\end{figure*}

\paragraph{Chain-of-thought, rationale supervision, and distillation.}
Scratchpad and chain-of-thought (CoT) methods prompt or train models to expose intermediate computation before producing an answer~\citep{nye2021show,wei2022chain}.
Subsequent work improves this paradigm through zero-shot prompting, self-consistency, rationale bootstrapping, synthetic supervision, and distillation~\citep{kojima2022large,wang2023selfconsistency,zelikman2022star,hsieh2023distilling,magister-etal-2023-teaching}.
These studies show that intermediate text and teacher-generated rationales can improve reasoning, but most supervised targets remain linear rationales followed by final answers.

\paragraph{Decomposition and structured inference.}
Decomposition-based methods break complex problems into subproblems, search structures, tool calls, or executable programs~\citep{zhou2023least,press2023measuring,khot2023decomposed,yao2023tree,besta2024graph,yao2023react,gao2023pal}.
They provide structured reasoning procedures at inference time, often relying on prompting strategies, external tools, or explicit exploration over intermediate states.

\paragraph{Process supervision and rationale faithfulness.}
Verifier and process-supervision methods train evaluators or reward models for complete solutions or intermediate reasoning steps~\citep{cobbe2021training,uesato2022solving,lightman2023lets}.
Related work on rationales and explainable NLP studies evidence selection, rationale quality, and explanation faithfulness~\citep{lei2016rationalizing,deyoung2020eraser,jain2019attention,turpin2023language,lanham2023measuring}.
Prior work motivates supervision while cautioning against interpreting intermediate text as faithful reasoning; accordingly, \dais{} evaluates whether dependency-conditioned supervision improves final-answer accuracy without claiming faithful intermediate explanations.

\section{Method}
\label{sec:method}

We introduce \dais{}, a dependency-aware intermediate supervision framework for constructing training data from teacher-generated CoT rationales.
As illustrated in Figure~\ref{fig:methods}, \dais{} collects teacher rationales, decomposes them into local subtasks, induces a dataset-level subtask schema, and converts each instance into a set of stage-level QA records.

\subsection{Problem Setup}
\label{sec:problem_setup}

We consider supervised reasoning tasks with training instances $(x_i,c_i,y_i)$, where $x_i$ is the task input, $c_i$ is optional context or evidence, and $y_i$ is the gold final answer.
For each training instance, a strong teacher model generates a CoT rationale $r_i$.
The teacher rationale is used only for constructing supervision targets and is not available at test time.

A standard CoT SFT baseline trains on a flat rationale-answer target:
\begin{equation}
\tau_i^{\mathrm{CoT}} = r_i \oplus y_i,
\end{equation}
where $\oplus$ denotes sequence concatenation.
This target exposes intermediate reasoning text, but treats the rationale as a single sequence and does not explicitly represent local subproblems or dependencies among intermediate states.

\dais{} instead constructs an ordered set of intermediate QA states
\[
T_i^{\mathrm{DAIS}}=[(q_{i1},a_{i1}),\ldots,(q_{ik_i},a_{ik_i})].
\]
These states are materialized as stage-level SFT records: the first record predicts $a_{i1}$ from $(x_i,c_i,q_{i1})$, each later record predicts $a_{it}$ from $(x_i,c_i,q_{it},a_{i,<t})$, and an additional final record predicts $y_i$ from $(x_i,c_i)$ alone.

At test time, the model receives only $x_i$ and optional $c_i$; no teacher rationale, gold intermediate state, or external decomposition module is provided.
Thus, \dais{} uses dependencies to organize SFT records, not as an inference-time requirement.

\subsection{Teacher CoT Collection}
\label{sec:teacher_cot_collection}

The first stage of \dais{} collects teacher rationales for the training instances.
For each example $(x_i,c_i,y_i)$, we prompt a strong teacher model with the task input, optional context, and answer options.
The teacher generates a step-by-step rationale $r_i$ together with a predicted final answer $\hat{y}_i$.

To reduce noise in teacher-generated rationales, we apply filtering before downstream construction.
For each instance, the teacher may generate up to $K$ candidate rationales.
We keep the first candidate whose final answer matches the gold answer $y_i$ and whose rationale is internally consistent: its substantive reasoning steps should be grounded in the input, context, answer options, or preceding steps, and should not support a different conclusion.
If no candidate passes the filter, the instance is discarded.
The retained rationales are treated as silver supervision for subsequent subtask decomposition and dependency-aware QA construction.

\subsection{Subtask Decomposition}
\label{sec:subtask_decomposition}

After collecting filtered teacher rationales, \dais{} decomposes each rationale into a compact ordered sequence of local subtasks.
Rather than splitting the CoT at the sentence level, we extract key reasoning operations that are necessary for reaching the final answer or supporting later intermediate decisions.
For each teacher rationale $r_i$, we extract
\begin{equation}
r_i
\rightarrow
S_i
=
\big((s_{i1},e_{i1}),\ldots,(s_{ik_i},e_{ik_i})\big),
\end{equation}
where $s_{it}$ denotes an instance-specific subtask and $e_{it}$ is its aligned teacher-rationale span.
In practice, we retain two to five main subtasks for most rationales, which keeps the supervision compact while preserving the main reasoning operations.

A valid subtask should be decision-relevant, locally answerable from the input, optional context, and aligned rationale span, and focused on one distinct reasoning operation rather than a stylistic transition or question restatement.
Adjacent spans are merged when they perform the same operation, while spans that mix separable operations are split.
The resulting instance-specific subtask candidates are normalized into a compact dataset-level schema and later converted into stage-level QA records.

\subsection{Dataset-level Subtask Schema Induction}
\label{sec:schema_induction}

\begin{figure}[t]
    \centering
    \includegraphics[width=\linewidth]{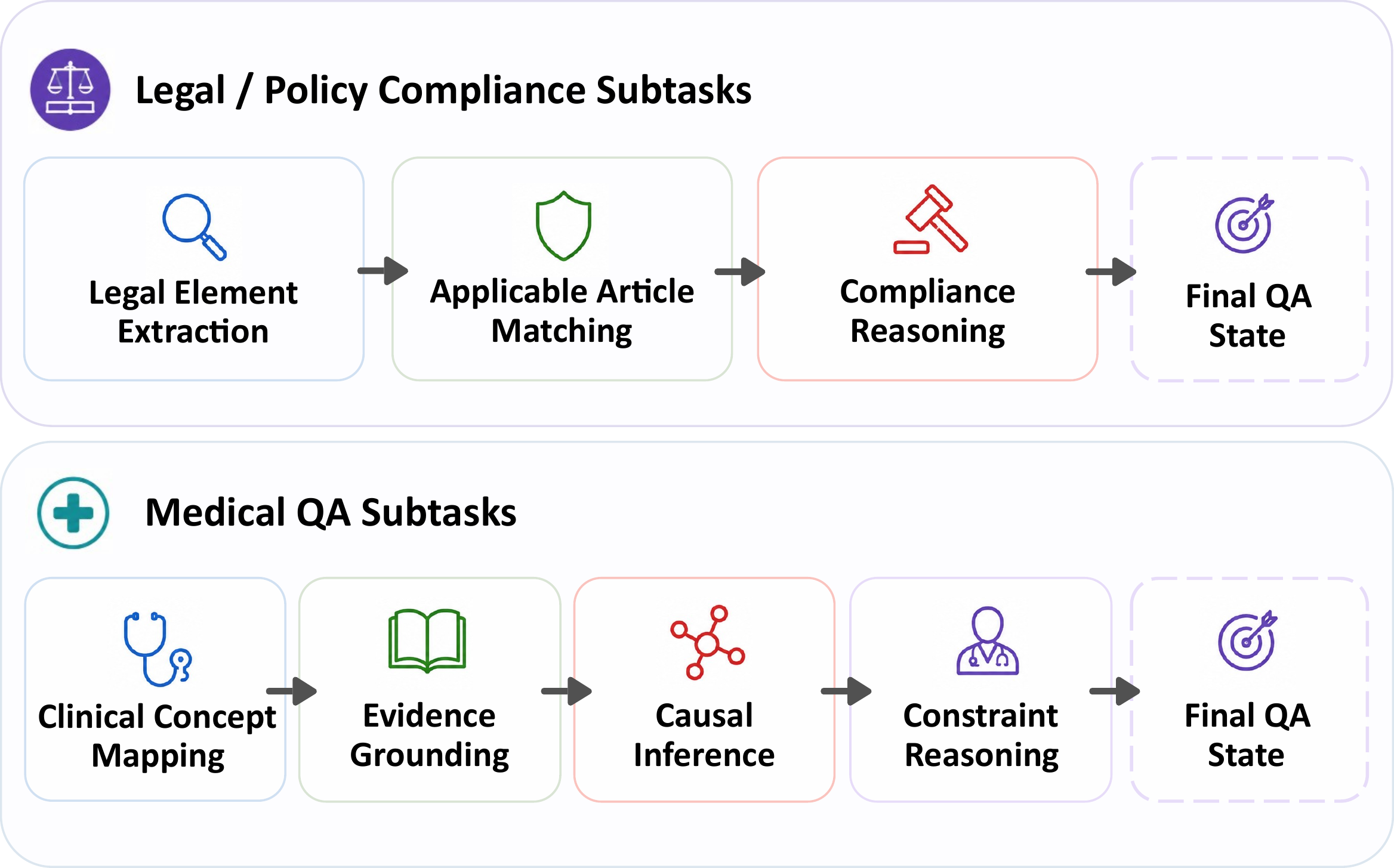}
    \caption{
    Examples of task-conditioned subtask schemas.
    }
    \label{fig:subtasks}
\end{figure}

The subtasks extracted from teacher rationales are initially instance-specific and free-form.
Even when two subtasks serve a similar reasoning function, the teacher model may describe them with different names, granularities, or surface forms.
To reduce this inconsistency, \dais{} performs dataset-level subtask schema induction using only the training rationales.

We collect all extracted subtasks in a dataset as a candidate pool and use their frequencies to guide the selection of high-coverage types.
We then induce a compact schema $\mathcal{Z}=\{z_1,\ldots,z_M\}$ by grouping candidates that share similar underlying reasoning functions rather than merely similar surface forms.
Each instance-level subtask is mapped to one schema type:
\begin{equation}
z_{it} = g(s_{it}), \quad z_{it}\in\mathcal{Z}.
\end{equation}
When a task has a stable domain-specific reasoning schema, such as legal or policy-compliance reasoning, we instantiate $\mathcal{Z}$ directly.
Otherwise, the schema is obtained through an LLM-assisted clustering and refinement process, with consistency checks used to merge overlapping types and revise inconsistent assignments.

The induced schema is task-conditioned rather than universal; its labels organize local-question templates and support analysis rather than general reasoning primitives.
Figure~\ref{fig:subtasks} shows example schemas, while each instance retains its own content and subtask order.

\subsection{DAIS QA Construction}
\label{sec:dependency_iqa}

After schema induction, \dais{} rewrites each selected subtask $s_{it}$ into a local QA state $(q_{it}, a_{it})$ and materializes it as an instruction--input--output record.
The first subtask record uses the original task input, optional context, and the current subtask question.
Each later subtask record additionally includes a serialized previous-state block containing selected earlier subtask questions and answers that are used as support context for the current local decision.
The output of each intermediate record is the local answer $a_{it}$.
The final-answer record is kept separate from these dependent subtask records.
Its input contains only the original task input and optional context, and its output is the gold final answer $y_i$.
Thus, \dais{} preserves the original final-task format while adding dependency-conditioned subtask supervision.
In our implementation, dependency conditioning is operationalized through previous-state serialization.
We do not train the model to predict explicit dependency-edge tokens, and the schema labels are used only to organize local QA templates.

The intermediate answer $a_{it}$ is derived from the aligned teacher-rationale span $e_{it}$, either by direct extraction or by meaning-preserving normalization.
During construction, we enforce three constraints: previous-state conditioning must be causal; each local answer must be grounded in the original input, optional context, answer options, or aligned rationale span; and the current subtask input must not reveal its own answer $a_{it}$ or the final answer $y_i$.

\subsection{Dataset Construction and Usage}
\label{sec:sft_dataset_usage}

The final stage converts each retained trace into instruction--input--output records.
For a retained instance with $k_i$ intermediate subtasks, \dais{} creates $k_i$ dependent subtask records and one plain final-answer record:
\begin{equation}
\mathcal{R}_i^{\mathrm{DAIS}}
=
\{(I_{it},a_{it})\}_{t=1}^{k_i}
\cup
\{(I_{iF},y_i)\}.
\end{equation}
Here $I_{it}$ is the input for the $t$-th subtask record, which contains the original input, the current subtask question, and the serialized previous-state context $H_{it}$ when $t>1$.
The final-answer record is defined by $I_{iF}=\operatorname{Format}(x_i,c_i)$ and $O_{iF}=y_i$, so it uses only the original input and optional context.

\dais{} does not introduce a new training algorithm or model architecture.
All records are trained with the standard autoregressive instruction-tuning objective.
Following the usual instruction-tuning setup, the loss is applied only to the output side of each instruction--input--output record, while the input side is treated as conditioning context.
For intermediate subtask records, the output is the local answer $a_{it}$; for the final-task record, the output is the gold final answer $y_i$.

\section{Experimental Setup}
\label{sec:experimental_setup}


\subsection{Tasks and Datasets}
\label{sec:tasks_datasets}

We use four complex reasoning benchmarks.
GDPR and AIACT evaluate privacy and policy-compliance reasoning \citep{li2025privaci,hu2025context}.
MedQA evaluates medical question answering \citep{jin2020disease}, and FOLIO evaluates natural-language logical reasoning \citep{han2022folio}.
All tasks are evaluated by final-answer accuracy.

\subsection{Compared Methods}
\label{sec:baselines}

We compare DAIS with target-format baselines under the same backbone and training data.
Base denotes the untuned backbone.
Final-SFT uses only final-answer supervision, while CoT-SFT trains on a flat teacher rationale followed by the final answer.
IndepQA uses the same intermediate QA states as DAIS but removes previous-state conditioning, isolating the effect of previous-state support context.
DAIS uses the full dependency-conditioned intermediate QA target.

For GDPR and AIACT, we additionally report CR-Data SFT, our Qwen-based model fine-tuned on the public Context-Reasoner SFT data released by \citet{li2025privaci}.
Because it uses an external data source, we treat it as a reference comparison rather than a controlled target-format ablation.

\begin{table*}[t]
\centering
\small
\setlength{\tabcolsep}{9.5pt}
\renewcommand{\arraystretch}{1.03}
\caption{
Accuracy on four benchmarks across four Qwen backbones.
\textsc{IndepQA} removes previous-state conditioning, and \textsc{CR-Data SFT} is an external-data reference for policy tasks.
Bold marks the best score in each column; averages are over backbones.
}\label{tab:main_results}
\vspace{-1pt}
\begin{tabular}{clccccc}
\toprule
\textbf{Benchmark} 
& \textbf{Method} 
& \textbf{Qwen2.5-3B} 
& \textbf{Qwen2.5-7B} 
& \textbf{Qwen3-4B} 
& \textbf{Qwen3-8B} 
& \textbf{Average} \\
\midrule
\multirow{6}{*}{GDPR}
& \textsc{Base}        & 0.756 & 0.890 & 0.608 & 0.768 & 0.756 \\
& \textsc{Final-SFT}   & 0.788 & 0.904 & 0.626 & 0.770 & 0.772 \\
& \textsc{CoT-SFT}     & 0.752 & 0.896 & 0.630 & 0.770 & 0.762 \\
& \textsc{IndepQA}     & 0.686 & 0.852 & 0.568 & 0.750 & 0.714 \\
& \textsc{CR-Data SFT} & 0.724	& 0.912	& 0.652	& 0.712 & 0.750  \\
\cmidrule{2-7}
& \textsc{DAIS (Ours)}       & \textbf{0.798} & \textbf{0.954} & \textbf{0.656} & \textbf{0.786} & \textbf{0.799} \\
\midrule
\multirow{6}{*}{AIACT}
& \textsc{Base}        & 0.356 & 0.408 & 0.628 & 0.700 & 0.523 \\
& \textsc{Final-SFT}   & 0.364 & 0.404 & 0.632 & 0.724 & 0.531 \\
& \textsc{CoT-SFT}     & 0.360 & 0.408 & 0.636 & 0.726 & 0.533 \\
& \textsc{IndepQA}     & 0.352 & 0.418 & 0.574 & 0.612 & 0.489 \\
& \textsc{CR-Data SFT} & 0.396 &0.382 &  0.654	& 0.692 & 0.531 \\
\cmidrule{2-7}
& \textsc{DAIS (Ours)}       & \textbf{0.472} & \textbf{0.486} & \textbf{0.658} & \textbf{0.738} & \textbf{0.589} \\
\midrule
\multirow{5}{*}{MedQA}
& \textsc{Base}        & 0.752 & 0.850 & 0.602 & 0.855 & 0.765 \\
& \textsc{Final-SFT}   & 0.738 & 0.845 & 0.732 & 0.848 & 0.791 \\
& \textsc{CoT-SFT}     & 0.752 & 0.849 & 0.758 & \textbf{0.858} & 0.804 \\
& \textsc{IndepQA}     & 0.722 & 0.818 & 0.750 & 0.850 & 0.785 \\
\cmidrule{2-7}
& \textsc{DAIS (Ours)}       & \textbf{0.764} & \textbf{0.872} & \textbf{0.769} & 0.856 & \textbf{0.815} \\
\midrule
\multirow{5}{*}{FOLIO}
& \textsc{Base}        & 0.470 & 0.560 & 0.710 & 0.835 & 0.644 \\
& \textsc{Final-SFT}   & 0.475 & 0.565 & 0.795 & 0.795 & 0.658 \\
& \textsc{CoT-SFT}     & 0.485 & 0.570 & 0.795 & 0.805 & 0.664 \\
& \textsc{IndepQA}     & 0.473 & 0.565 & 0.785 & 0.795 & 0.655 \\
\cmidrule{2-7}
& \textsc{DAIS (Ours)}       & \textbf{0.500} & \textbf{0.580} & \textbf{0.815} & \textbf{0.840} & \textbf{0.684} \\
\bottomrule
\end{tabular}
\end{table*}

\subsection{Implementation and Evaluation}
\label{sec:implementation_eval}
Intermediate QA records are constructed only from training-set rationales generated by DeepSeek models \citep{deepseekai2026deepseekv4}. 
For GDPR and AIACT, we use a predefined compliance-oriented schema; for MedQA and FOLIO, we induce dataset-level schemas with LLM-assisted clustering and manual audit. 
In \dais{}, later subtask records receive serialized previous states as context, while the final-answer record remains the original task. 
All supervised variants for the same dataset and backbone use the same retained training instances. 
If a rationale fails consistency or answer-matching checks, the corresponding instance is removed for all variants. 
More details are provided in Appendix~\ref{app:trace_construction_details}.

Controlled experiments fine-tune four Qwen backbones: Qwen2.5-3B, Qwen2.5-7B \citep{qwen2025qwen25technicalreport}, Qwen3-4B, and Qwen3-8B \citep{yang2025qwen3technicalreport}.
Reference comparisons additionally include Qwen2.5-72B, GPT-OSS-120B \citep{agarwal2025gpt}, MiniMax-M27 \citep{minimax2026m27}, GLM-4.7 \citep{5team2025glm45agenticreasoningcoding}, and released Context-Reasoner SFT/RL models \citep{li2025privaci}.

At evaluation time, models receive only the original task input and optional context.
The main evaluation uses the final-answer task format.
Final predictions are normalized with a fixed LLM-based answer matcher under the same task-specific rubric across methods.
The matcher maps each model output to a canonical option or label and compares it with the gold benchmark answer; ambiguous outputs, missing answers, or multiple incompatible final answers are counted as incorrect.
The matcher is used for answer extraction and normalization, not for creating test labels.

\section{Experimental Results}
\label{sec:results}

In this section, we evaluate overall accuracy, dependency-validity ablations, subtask coverage, component variants, and efficiency with respect to original training instances.

\subsection{Main Results}
\label{sec:main_results}
\begin{table}[t]
\centering
\small
\setlength{\tabcolsep}{4.5pt}
\renewcommand{\arraystretch}{1.03}
\caption{
Accuracy comparison on GDPR and AI Act benchmarks for 7B variants, larger reference models, and context-reasoner systems.
}
\label{tab:large_reference}
\vspace{-1pt}
\begin{tabularx}{\columnwidth}{
@{}
>{\centering\arraybackslash}p{0.21\columnwidth}
>{\raggedright\arraybackslash}p{0.35\columnwidth}
>{\centering\arraybackslash}p{0.15\columnwidth}
>{\centering\arraybackslash}p{0.15\columnwidth}
@{}
}
\toprule
\textbf{Group} & \textbf{Model} & \textbf{GDPR} & \textbf{AI Act} \\
\midrule

\multirow[c]{3}{=}{\centering\footnotesize\itshape Qwen-7B\\variants}
& Qwen2.5-7B (Base)          & 0.890 & 0.408 \\
& Qwen2.5-7B + CoT           & 0.896 & 0.408 \\
& \textbf{Qwen2.5-7B + DAIS} & \textbf{0.954} & 0.486 \\

\midrule
\multirow[c]{4}{=}{\centering\footnotesize\itshape Reference\\models}
& Qwen2.5-72B                & 0.756 & 0.610 \\
& GLM-4.7                    & 0.842 & 0.610 \\
& MiniMax-M27                & 0.906 & 0.616 \\
& GPT-OSS-120B               & 0.848 & 0.678 \\

\midrule

\multirow[c]{2}{=}{\centering\footnotesize\itshape CtxR\\systems}
& CtxR-SFT-7B                & 0.820 & 0.764 \\
& CtxR-RL-7B                 & 0.822 & \textbf{0.780} \\

\bottomrule
\end{tabularx}
\end{table}

Table~\ref{tab:main_results} reports the controlled comparison across four benchmarks and four backbone models.
By point estimate, \dais{} achieves the best average over backbones on all four tasks and obtains the best result in 15 of 16 benchmark--model settings.
The advantage is most evident on policy-oriented benchmarks, especially AIACT: averaged over backbones, \dais{} improves over the strongest non-\dais{} baseline by 2.7 points on GDPR and 5.6 points on AIACT.
This suggests that dependency-conditioned subtask supervision is particularly helpful when final decisions require linking scenario facts to applicable constraints and intermediate conditions.

\dais{} also improves the average accuracy on MedQA and FOLIO, although the margins are smaller and one MedQA setting is a near-tie: on Qwen3-8B MedQA, \textsc{CoT-SFT} reaches 0.858, while \dais{} obtains 0.856.
On the policy benchmarks, \dais{} remains stronger than \textsc{CR-Data SFT}; on AIACT, for example, the average accuracy increases from 0.531 to 0.589.
Together with its consistent advantage over \textsc{IndepQA}, this suggests that dependency preservation is an important contributor beyond QA formatting alone.

Table~\ref{tab:large_reference} further shows that \dais{} achieves competitive GDPR performance relative to several larger reference models and CtxR systems, although specialized CtxR models remain substantially stronger on AIACT. These comparisons are intended as broad reference points rather than controlled head-to-head evaluations, since differences in prompting, decoding, evaluation protocols, and training setups may also affect performance.

\subsection{Dependency Validity}
\label{sec:dependency_validity}

Table~\ref{tab:dependency_validity} provides a representative dependency-validity diagnostic on AIACT and MedQA.
It examines whether valid previous-state conditioning is a useful training signal beyond surface properties of the SFT records.
We compare against final-only SFT (\textsc{Final}), a length-matched control (\textsc{Match}), order-marked subtask records (\textsc{Ordered}), corrupted previous states (\textsc{Corr.}), and dependency-free intermediate QA (\textsc{IndepQA}).
\textsc{Match} controls for target length at the original-instance level.
\textsc{Ordered} preserves subtask and order markers but removes previous-state conditioning.
\textsc{Corr.} preserves the dependency-style template but replaces previous-state fields with mismatched states.
\textsc{IndepQA} exposes the same intermediate QA states as independent records without previous states.

\begin{table}[t]
\centering
\small
\renewcommand{\arraystretch}{1.03}
\setlength{\tabcolsep}{4.2pt}
\caption{
Accuracy on dependency validity ablations.
Corr. preserves the dependency format but uses mismatched previous states.
}
\vspace{-1pt}
\begin{tabular}{lcc|cc}
\toprule

& \multicolumn{2}{c|}{\textbf{AIACT}}
& \multicolumn{2}{c}{\textbf{MedQA}} \\

\cmidrule(r){2-3}
\cmidrule(l){4-5}

\textbf{Method}
& \textbf{Qwen-7B}
& \textbf{Qwen-4B}
& \textbf{Qwen-7B}
& \textbf{Qwen-4B} \\

\midrule

Final & 0.404 & 0.632 & 0.845 & 0.732 \\

Match & 0.440 & 0.622 & 0.814 & 0.766 \\

Ordered & 0.382 & 0.504 & 0.820 & 0.758 \\

Corr.
& 0.400
& 0.568
& 0.812
& 0.754 \\

IndepQA
& 0.418
& 0.534
& 0.818
& 0.574 \\
\midrule
\textbf{DAIS}
& \textbf{0.486}
& \textbf{0.638}
& \textbf{0.872}
& \textbf{0.769} \\

\bottomrule
\end{tabular}
\label{tab:dependency_validity}
\end{table}

\dais{} achieves the best accuracy across all four combinations of dataset and backbone.
On AIACT, it exceeds \textsc{Match} by 4.6 points and \textsc{Corr.} by 8.6 points on Qwen2.5-7B, and also gives the strongest result on Qwen3-4B.
On MedQA, \dais{} remains best for both backbones, although the margin over \textsc{Match} is small on Qwen3-4B.
Because \textsc{Corr.} preserves the dependency-style format while corrupting the referenced previous states, its gap from \dais{} indicates that valid support relations among intermediate states are an important part of the observed gains.
These ablations address the central training-time question in \dais{}: whether valid previous-state conditioning improves the auxiliary supervision signal.
They should not be read as evidence that the model executes the same dependency graph at inference time, because the main evaluation supplies only the original task input and context.
Rather, they show that preserving valid support relations during SFT matters beyond target length, ordering cues, dependency-like templates, or intermediate QA exposure alone.

\subsection{Subtask Coverage}
\label{sec:subtask_coverage}

Table~\ref{tab:subtask_coverage} examines whether exposing only partial subtask information is sufficient to reproduce the gains of \dais{}.
All variants use the same original-task final-answer records.
The subtask-only variants add records from only one subtask subset, whereas \dais{} adds all dependency-conditioned subtask records.
We also compare with \textsc{IndepQA}, which exposes intermediate QA states without previous-state conditioning.

\begin{table}[t]
\centering
\small
\renewcommand{\arraystretch}{1.03}
\setlength{\tabcolsep}{3.8pt}
\caption{
Accuracy of subtask coverage variants.
\textsc{DAIS} preserves the full dependency organization, while the other variants expose only partial or dependency free intermediate states.
}
\vspace{-1pt}
\begin{tabular}{lcc|cc}
\toprule
& \multicolumn{2}{c|}{\textbf{GDPR}}
& \multicolumn{2}{c}{\textbf{MedQA}} \\
\cmidrule(lr){2-3}\cmidrule(lr){4-5}
\textbf{Variant}
& \textbf{Qwen-7B}
& \textbf{Qwen-4B}
& \textbf{Qwen-7B}
& \textbf{Qwen-4B} \\
\midrule
Base     & 0.890 & 0.608 & 0.850 & 0.602 \\
Subtask1 & 0.896 & 0.622 & 0.810 & 0.754 \\
Subtask2 & 0.899 & 0.618 & 0.812 & 0.746 \\
Subtask3 & 0.897 & 0.604 & 0.814 & 0.754 \\
IndepQA  & 0.852 & 0.568 & 0.818 & 0.750 \\
\midrule
\textbf{DAIS}
         & \textbf{0.954} & \textbf{0.656}
         & \textbf{0.872} & \textbf{0.769} \\
\bottomrule
\end{tabular}
\label{tab:subtask_coverage}
\end{table}

\dais{} achieves the best accuracy in all four reported settings.
On GDPR with Qwen2.5-7B, the individual subtask variants range from 0.896 to 0.899, while \dais{} reaches 0.954.
On MedQA, adding only one subtask subset helps Qwen3-4B but remains below \dais{}; for Qwen2.5-7B, these variants are below the base model.
These results suggest that partial subtask exposure can provide signals in some cases, but the gain appears to come not only from revealing intermediate subtask content; connecting and aggregating those states through valid dependencies also matters.

\subsection{Related-Style Component Variants}
\label{sec:single_component}

Table~\ref{tab:single_component} evaluates controlled component-isolation variants inspired by prior decomposition and rationale-supervision methods.
These variants are not full reproductions of prior systems; instead, they isolate supervision components often emphasized in such work, including high-level decomposition plans, subproblem questions, and intermediate answers.
Using the same original instances and Qwen2.5-7B backbone, all variants keep the original-task final-answer records fixed and modify only the additional \dais{}-derived records.
\textsc{p-only} retains only the overall plan, represented as the ordered list of subtask names for the instance.
\textsc{a-only} retains only the intermediate answer components.
\textsc{q-only} retains only the local subtask questions.
We compare these variants with standard baselines and the full \dais{} construction.

\begin{table}[t]
\centering
\small
\renewcommand{\arraystretch}{1.03}
\setlength{\tabcolsep}{7pt}
\caption{
Accuracy of single-component variants on Qwen2.5-7B.
}
\vspace{-1pt}
\begin{tabular}{lcccc}
\toprule
\textbf{Method}
& \textbf{GDPR}
& \textbf{AIACT}
& \textbf{MedQA}
& \textbf{FOLIO} \\
\midrule
Base      & 0.890 & 0.408 & 0.850 & 0.560 \\
FINAL       & 0.904 & 0.404 & 0.845 & 0.565 \\
CoT-SFT     & 0.896 & 0.408 & 0.849 & 0.570 \\
\midrule
P-only    & 0.906 & 0.412 & 0.794 & 0.550 \\
A-only    & 0.902 & 0.406 & 0.794 & 0.545 \\
Q-only    & 0.896 & 0.410 & 0.798 & 0.555 \\
\midrule
\textbf{DAIS}
          & \textbf{0.954}
          & \textbf{0.486}
          & \textbf{0.872}
          & \textbf{0.580} \\
\bottomrule
\end{tabular}

\label{tab:single_component}
\end{table}

The component variants remain below full \dais{}, supporting the value of coupling local QA signals with valid previous-state conditioning.

\subsection{Data Efficiency}
\label{sec:data_efficiency}

\begin{figure}[t]
    \centering
    \includegraphics[width=\linewidth]{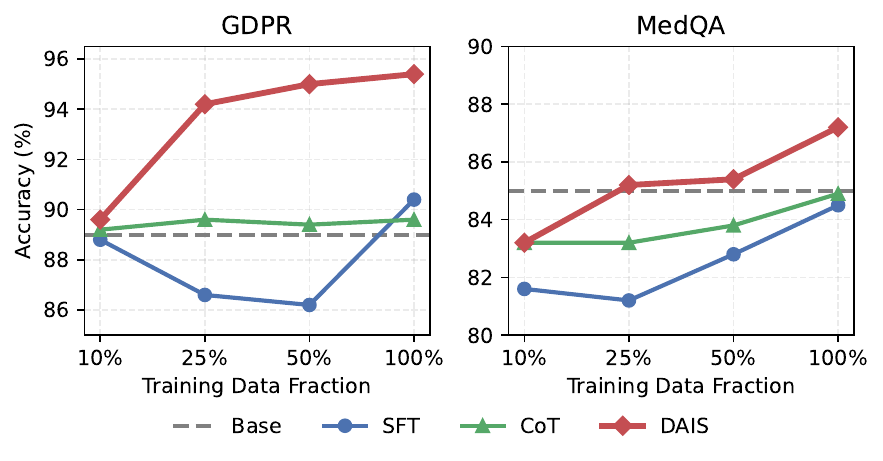}
    \caption{
Sample efficiency on GDPR and MedQA with Qwen2.5-7B.
The dashed line denotes the untuned base model.
}
    \label{fig:efficiency}
\end{figure}

Figure~\ref{fig:efficiency} studies data efficiency with 10\%, 25\%, 50\%, and 100\% of the original supervised instances.
We compare \dais{} with SFT and CoT baselines on Qwen2.5-7B.
For \dais{}, each sampled instance contributes its dependency-conditioned subtask records and the original-task final-answer record.

On GDPR, \dais{} is strongest at every fraction, reaching 0.896 with 10\% of the original instances and 0.942 with 25\%.
On MedQA, the gains are more gradual: \dais{} surpasses the base model from 25\% onward and achieves the best full-data result.
These results suggest improved efficiency with respect to original supervised instances, especially on GDPR.
However, because \dais{} expands each instance into multiple records and additional supervision tokens, the comparison does not control for token budget, optimization steps, or training compute.

\section{Discussion}
\label{sec:discussion}

Ablations against \textsc{IndepQA}, \textsc{Ordered}, \textsc{Match}, and \textsc{Corr.} indicate that \dais{} is not explained only by target length, ordering cues, dependency-like templates, or intermediate QA exposure.
The results support valid previous-state conditioning as a useful training-time supervision signal.
Because the main evaluation uses direct final-answer inference without teacher rationales, gold intermediate states, or external decomposition modules, these findings demonstrate improved SFT target organization and final-answer accuracy, but do not establish that the model faithfully executes an explicit dependency graph internally.

\dais{} also introduces additional target construction and expands each retained instance into multiple SFT records.
At the 1,000-example scale, the preprocessing overhead on MedQA remained moderate: CoT generation took about 1.7 hours and subtask decomposition about 4.6 hours.
The low-resource curves further suggest that fewer original supervised instances may suffice in domains such as GDPR.
However, these comparisons evaluate efficiency with respect to original instances rather than matched token budget, record count, wall-clock training time, or total training cost.

\section{Conclusion}

In this work, we introduced Dependency-Aware Intermediate QA Supervision (DAIS), a framework that integrates dependency-conditioned intermediate QA into supervised fine-tuning.
Through experiments across policy compliance, medical QA, and logical reasoning benchmarks, we show that DAIS consistently improves final-answer accuracy over answer-only, flat CoT, and independent-QA baselines.
Our results indicate that intermediate text alone is insufficient: effective supervision should also specify how earlier local conclusions support later decisions.
Ablations confirm that valid previous-state conditioning contributes beyond target length, ordering cues, and QA formatting.
Future work may further examine inference-time faithfulness and more scalable construction of dependency-aware supervision.

\section*{Limitations}

\paragraph{Generated supervision artifacts.}
\dais{} depends on generated or automatically constructed supervision artifacts, including teacher-generated rationales or CoT traces, intermediate QA states, subtask decompositions, role labels, and dependency links.
These artifacts may contain hallucinated evidence, unsupported subtasks, incomplete intermediate states, mislabeled roles, or incorrect dependency relations.
They are also not guaranteed to be optimal or faithful descriptions of the model's internal computation.
Stronger teachers, better prompts, human annotation, filtering, or iterative refinement may produce higher-quality dependency-conditioned targets.

\paragraph{Filtering and retained-set bias.}
The filtering step changes the effective training distribution.
Although all controlled variants use the same retained instances, so target-format comparisons remain controlled, the retained set may be biased toward examples for which the teacher can produce gold-consistent rationales.
Future work should report retention rates, label-wise retention, and the properties of discarded examples.

\paragraph{SFT-only setting.}
Our study focuses on supervised fine-tuning in order to isolate the effect of dependency-conditioned target organization.
This controlled setting allows us to compare final-only supervision, flat CoT supervision, independent intermediate QA, corrupted dependency supervision, and dependency-conditioned supervision under the same likelihood objective.
However, it also limits the scope of our conclusions.
We do not combine \dais{} with reinforcement learning, preference optimization, rejection sampling, self-training, inference-time search, or verifier-guided refinement.
These methods may further improve performance, but they introduce additional factors beyond the controlled SFT comparison studied here.

\paragraph{Main-inference setting and structured-inference extension.}
In the main experiments, \dais{} does not add inference-time inputs or decoding steps: the model receives only the original task input and optional context.
The additional cost is primarily in target construction and SFT data expansion.
A different deployment mode could ask the model to generate \dais{}-style intermediate states at test time and then condition later predictions on generated states.
Such structured inference would introduce additional latency and error-propagation risks, because early-state mistakes could affect later local decisions.
We do not evaluate this deployment setting in the main results.

\paragraph{Evaluation scope.}
Our evaluation is limited to selected policy-compliance, medical QA, multi-hop, and logical-reasoning benchmarks.
These experiments primarily measure final-answer accuracy and selected properties of generated intermediate traces.
They do not prove that intermediate QA states faithfully reflect the model's internal computation, nor do they guarantee that the same gains will hold in domains where teacher rationales, subtask decompositions, or dependency structures are less reliable.
The role labels used in our data are task-conditioned supervision markers rather than universal reasoning primitives.

\paragraph{Answer matching.}
We use a fixed LLM-based answer matcher to normalize model generations into canonical task labels or answer options when deterministic parsing is insufficient.
This matcher does not create benchmark labels, but it may introduce evaluator-model bias or prompt sensitivity.
We therefore use the same matcher and rubric across methods and count ambiguous outputs as incorrect.

\section*{Ethical Considerations}

\dais{} is evaluated on tasks involving policy compliance, legal or regulatory reasoning, medical QA, multi-hop QA, and logical reasoning.
Models trained with this framework should be used for research analysis or decision support rather than as substitutes for qualified legal, regulatory, or medical experts.
Structured intermediate states may appear coherent even when they contain unsupported, incomplete, or incorrect reasoning.

The supervision artifacts used by \dais{} are constructed with the help of large language models, including rationales, intermediate QA states, role labels, and dependency links.
These artifacts may inherit hallucinations, biases, or unsupported assumptions from the teacher model or generation pipeline.
Before public release or downstream use, generated supervision data should be audited and documented, including provenance, prompting procedures, filtering criteria, known noise sources, and intended-use restrictions.

We use a fixed LLM-based answer matcher for final-answer extraction and normalization, and may also use LLMs for auxiliary analyses such as checking intermediate-state quality or categorizing error types.
The matcher does not create benchmark labels; it maps model outputs to canonical task labels or answer options and compares them with the gold answers.
Nevertheless, LLM-based matching can reflect evaluator-model biases, rubric sensitivity, or prompt sensitivity.
We therefore use the same matcher, decoding setting, and rubric across methods, count ambiguous or incompatible outputs as incorrect, and treat such evaluation as a limitation of the current study.

Finally, dependency-aware intermediate traces should not be interpreted as certified explanations of model internals.
They are supervision and analysis artifacts designed to improve and diagnose model behavior, not guarantees of faithful reasoning.
Users should avoid over-trusting a model's output solely because it provides a structured intermediate trace.

\bibliography{custom}

\clearpage
\appendix

\section{Experimental Details}
\label{app:experimental_details}

This section provides additional implementation details for data construction, fine-tuning, model deployment, and evaluation.
Dataset sources and licenses are summarized in Table~\ref{tab:datasources}, and model sources and licenses are summarized in Table~\ref{tab:modelsources}.
All construction models are used only to build supervised training targets.
At evaluation time, no teacher rationale, gold intermediate state, or external decomposition module is provided.

\begin{table*}[t]
\centering
\small
\renewcommand{\arraystretch}{1.6}
\caption{Datasets, sources and licenses used in this work.} \label{tab:datasources}
\begin{tabularx}{0.95\textwidth}{l|l}
\toprule
\textbf{Dataset} & \textbf{URL}  \\ \midrule
PrivaCI-Bench-gdpr & https://github.com/HKUST-KnowComp/PrivaCI-Bench  \\
PrivaCI-Bench-AIACT & https://github.com/HKUST-KnowComp/PrivaCI-Bench  \\
PrivaCI-Bench-sft dataset & https://github.com/HKUST-KnowComp/PrivaCI-Bench \\
folio & https://github.com/Yale-LILY/FOLIO \\
medqa &  https://github.com/jind11/MedQA \\

\bottomrule
\end{tabularx}
\end{table*}

\subsection{Datasets}
\label{app:datasets}

We evaluate on four reasoning datasets covering legal or policy-compliance reasoning, medical question answering, and natural-language logical reasoning.
Dataset sources are summarized in Table~\ref{tab:datasources}.

For the privacy and policy-compliance domain, we use the GDPR and AIACT subsets from Privaci-Bench.
Privaci-Bench also contains a HIPAA subset, but we do not include it in our main experiments because it contained only slightly more than 200 examples at the time of our study.
This size was too small for constructing a label-balanced test set and a controlled fine-tuning split comparable to the GDPR and AIACT settings.
We therefore focus on GDPR and AIACT, which provide sufficiently large compliance-reasoning pools and cover two related but distinct regulatory domains.
For GDPR, the original pool contains 3,137 examples; for AIACT, the original pool contains 3,000 examples.

The GDPR dataset evaluates whether a described event is permitted, prohibited, or unrelated under GDPR-style privacy compliance.
The task requires identifying relevant actors, data types, processing purposes, and applicable regulatory obligations before producing the final norm judgment.
The AIACT dataset follows the same norm-judgment format, but focuses on AI-system compliance scenarios, such as AI-system roles, deployment contexts, biometric or high-risk use cases, and applicable policy constraints.
For both GDPR and AIACT, we randomly sample a label-balanced test set of 500 examples and then sample 1,000 training examples from the remaining pool.

MedQA is a medical multiple-choice question answering benchmark.
Each example contains a clinical question, answer options, and a gold answer.
Solving these questions often requires extracting clinical evidence, mapping symptoms or findings to medical concepts, eliminating incompatible options, and selecting the final diagnosis, treatment, or management decision.
We use 1,000 training examples and 500 test examples.

FOLIO is a natural-language logical reasoning benchmark with first-order-logic-style dependencies.
Each example contains natural-language premises, a conclusion, and a label indicating whether the conclusion is \texttt{True}, \texttt{False}, or \texttt{Unknown}.
The task requires retrieving relevant premises, composing local deductions, and determining the final veracity label.
We use 1,000 training examples and 200 test examples.

All sampled splits are fixed before trace construction and are shared across all target-format variants.
Teacher-rationale filtering and trace construction are applied only to the training split.
After filtering, all controlled supervised variants for the same dataset and backbone use the same retained training instances, original inputs, and gold final labels; they differ only in how the supervision is constructed.

\subsection{Trace Construction}
\label{app:trace_construction_details}

The intermediate QA records used for supervised fine-tuning are constructed through API calls.
For each training instance, we first generate a candidate teacher rationale with \texttt{DeepSeek-V4-Flash}.
The candidate is then validated by \texttt{DeepSeek-V4-Pro} for final-answer correctness and rationale consistency.
A rationale is considered invalid if its predicted answer does not match the gold answer, if it contains unsupported non-trivial reasoning steps, or if it includes explicit contradictions or support for an alternative answer.
If the initial candidate fails validation, we regenerate the rationale with \texttt{DeepSeek-V4-Pro} for up to two additional rounds and retain the first valid candidate.
All API calls used for trace construction are decoded with temperature $0.7$.

After obtaining a valid teacher rationale, we use \texttt{DeepSeek-V4-Pro} for subtask-level QA construction.
For GDPR and AIACT, we use a predefined compliance-oriented schema and generate local QA states for each schema stage, including legal element extraction, legal regulation mapping, and compliance reasoning.
For MedQA and FOLIO, we apply the full decomposition pipeline: \texttt{DeepSeek-V4-Pro} extracts two to five decision-relevant subtasks, aligns them with rationale spans where applicable, derives intermediate answers, and reconstructs local QA states.
We then induce dataset-level schemas for MedQA and FOLIO through LLM-assisted clustering with \texttt{DeepSeek-V4-Pro}, followed by manual auditing, merging, and relabeling.

For \dais{}, dependency conditioning is implemented by serializing previous subtask states into the input of later subtask records.
The first subtask record receives only the original task input, optional context, and the current subtask question.
Later subtask records additionally receive a \texttt{Previous states:} block containing earlier subtask questions, inputs, reasoning when available, and answers.
The final-answer record remains the original final-answer task and does not receive constructed intermediate states.
The \textsc{IndepQA} control uses the same local subtask records as \dais{} but removes the \texttt{Previous states:} context from later subtask inputs.

\subsection{Fine-tuning Setup}
\label{app:fine_tuning_setup}

All supervised fine-tuning experiments are conducted with LLaMA-Factory.
For controlled comparisons, all variants for the same task and backbone use the same retained training instances, original task inputs, gold final labels, optimization budget, and sequence-length limits.
They differ only in the supervised target construction.

\textsc{Final-SFT} uses one final-answer record per retained instance.
\textsc{CoT-SFT} uses one record whose output is the flat teacher rationale followed by the final answer.
\textsc{IndepQA} and \dais{} are both materialized as stage-level subtask records plus one final-answer record.
\textsc{IndepQA} uses the same local subtask records as \dais{} but removes the \texttt{Previous states:} context from later subtask inputs.
\dais{} keeps this previous-state context for later intermediate subtasks.
For both methods, the final-answer record uses only the original task input and gold final label.

All fine-tuned variants are optimized with the standard autoregressive SFT objective.
In instruction tuning, the loss is applied to the output side of each instruction--input--output record.
Training hyperparameters are held fixed across target-format variants for each backbone and task, so that the comparisons isolate the effect of the supervision format rather than changes in optimization.
Fine-tuning is performed with LLaMA-Factory on the mixed Ascend environment described in Appendix~\ref{app:model_deployment}, with device allocation determined by model size and memory requirements.

\subsection{Inference and Model Deployment}
\label{app:model_deployment}

\begin{table*}[t]
\centering
\small
\renewcommand{\arraystretch}{1.6}
\caption{Models, sources and licenses used in this work.} \label{tab:modelsources}
\begin{tabularx}{0.98\textwidth}{l|l|c}
\toprule
\textbf{Model} & \textbf{URL} & \textbf{Licenses} \\ \midrule
Qwen2.5-3B & https://huggingface.co/Qwen/Qwen2.5-3B-Instruct & Apache License 2.0 \\
Qwen2.5-7B & https://huggingface.co/Qwen/Qwen2.5-7B-Instruct & Apache License 2.0 \\
Qwen2.5-14B & https://huggingface.co/Qwen/Qwen2.5-14B-Instruct & Apache License 2.0 \\
Qwen2.5-72B & https://huggingface.co/Qwen/Qwen2.5-72B-Instruct & Qwen license \\
Qwen3-4B & https://huggingface.co/Qwen/Qwen3-4B & Apache License 2.0 \\
Qwen3-8B & https://huggingface.co/Qwen/Qwen3-8B & Apache License 2.0 \\
Qwen3-14B & https://huggingface.co/Qwen/Qwen3-14B & Apache License 2.0 \\
 GPT-OSS-120B & https://huggingface.co/openai/gpt-oss-120b & Apache License 2.0\\
 MiniMax-M27 & https://huggingface.co/MiniMaxAI/MiniMax-M2.7 & NON-COMMERCIAL LICENSE\\
GLM-4.7 & https://huggingface.co/zai-org/GLM-4.7 & MIT License\\
DeepSeek-V4-pro & https://www.deepseek.com/ & DEEPSEEK LICENSE\\
DeepSeek-V4-Flash & https://www.deepseek.com/ & DEEPSEEK LICENSE\\
context-reasoner-sft &https://huggingface.co/hubin/context-reasoner-sft\_open\_thinker & Qwen license\\
context-reasoner-rl & \renewcommand{\arraystretch}{1.05}\begin{tabular}[c]{@{}l@{}}https://huggingface.co/hubin/context-reasoner-ppo\_open\\ \_thinker\_acc\_reward\end{tabular} & Qwen license\\

\bottomrule
\end{tabularx}
\end{table*}

We use vLLM~\citep{kwon2023efficient} for local model inference and deployment.
The experiments are run on a mixed hardware environment consisting of 32GB Ascend 910B4 devices, 64GB Ascend 910B3 devices, and 32GB NVIDIA Tesla V100 PCIe devices.
We allocate devices according to model size and memory requirements.

For models with more than 70 billion parameters, we use eight 32GB Ascend 910B4 devices with tensor parallelism.
This configuration provides 256GB aggregate device memory before runtime overhead.
For 32B-scale models, including Qwen2.5-32B, Qwen3-32B, QwQ-32B, and DS-Qwen-32B, we use two 32GB Ascend 910B4 devices with tensor parallelism and keep the batch size and maximum generation length within the available memory budget.
For 14B-scale models, such as Qwen2.5-14B, we use one 64GB Ascend 910B3 device.
For 3B--9B-scale models, including LLaMA3.1-8B, Qwen2.5-7B, DS-Qwen-7B, and ChatGLM4-9B, we use one 32GB Ascend 910B4 device.
For retrieval-related components, such as \texttt{bge-m3} and \texttt{bge-reranker-v2-m3}, we use one 32GB NVIDIA Tesla V100 PCIe device.

The allocation follows the approximate bf16/fp16 parameter footprint of each model size before KV cache and framework overhead.
Accordingly, large models are deployed with tensor parallelism, and batch size and maximum generation length are kept within the available memory budget.

\subsection{Decoding and Evaluation}
\label{app:decoding_evaluation}

At evaluation time, each model receives only the original task input and optional context.
No teacher rationale, gold intermediate state, or external decomposition module is provided to the evaluated model.
The main evaluation uses the final-answer task format: the model is prompted to produce the final answer directly.
For methods or reference systems that produce rationales or structured text, the final prediction is extracted from the generated output using the same normalization procedure across methods.

We use a fixed LLM-based answer matcher to normalize candidate outputs.
For each test example, the matcher receives the original task input, optional context or answer options, the gold final answer, and the candidate model output.
It identifies the candidate's final prediction, maps it to the canonical task label or option when possible, and compares it with the gold answer.
Ambiguous outputs, missing answers, or outputs containing multiple incompatible final answers are counted as incorrect.
The same matcher model, evaluation prompt, deterministic decoding setting, and decision rubric are used across methods within each task.
The matcher is used for answer extraction and normalization rather than for creating benchmark labels.

We report final-answer accuracy for all tasks.
For multiple-choice tasks, a prediction is correct if the normalized final option or answer text matches the gold option.
For label-style tasks, a prediction is correct if the normalized final label matches the gold label, after accounting for task-specific verbalizations such as \texttt{permit}/\texttt{prohibit}/\texttt{unrelated} or \texttt{True}/\texttt{False}/\texttt{Unknown}.

For API-based trace construction, we use temperature $0.7$.
For local model generation, decoding settings are held fixed across methods within each task and backbone.
When evaluating generations that explicitly request long-form reasoning, we use stochastic decoding with temperature $1.0$ and top-$p=1.0$ to allow complete reasoning traces.
For other local model outputs, we use deterministic decoding with temperature $0$ and top-$p=1.0$.
The LLM-based answer matcher is run with deterministic decoding.
Each experiment is repeated three times, and all reported scores are averaged over the three runs.

\subsection{Reproducibility Notes}
\label{app:reproducibility_notes}

All controlled comparisons use the same sampled split and the same retained training instances for a given dataset and backbone.
Teacher-rationale filtering and trace construction are applied only to the training split.
After this step, all controlled supervised variants are trained on the same retained instances, original task inputs, and gold final labels; they differ only in the supervised target construction.
Training configuration, decoding settings, evaluation prompts, answer extraction, and evaluation scripts are kept fixed across target-format variants.
Dataset and model sources, together with license or terms-of-use information, are documented in Tables~\ref{tab:datasources} and~\ref{tab:modelsources}.

\section{Additional Experimental Results}
\label{app:additional_results}

\subsection{Reference Comparisons}
\label{app:reference_comparisons}

Table~\ref{tab:large_reference_appendix} reports additional reference comparisons on MedQA and FOLIO.
These results compare Qwen2.5-7B variants with larger reference models.
They are not controlled target-format comparisons, because the reference models differ in scale and training source.

\begin{table}[t]
\centering
\small
\setlength{\tabcolsep}{4.5pt}
\renewcommand{\arraystretch}{1.03}
\caption{
Accuracy comparison on MedQA and Folio benchmarks for 7B variants, larger reference models, and context-reasoner systems.
}
\label{tab:large_reference_appendix}

\begin{tabularx}{\columnwidth}{
@{}
>{\centering\arraybackslash}p{0.21\columnwidth}
>{\raggedright\arraybackslash}p{0.35\columnwidth}
>{\centering\arraybackslash}p{0.15\columnwidth}
>{\centering\arraybackslash}p{0.15\columnwidth}
@{}
}
\toprule
\textbf{Group} & \textbf{Model} & \textbf{MedQA} & \textbf{Folio} \\
\midrule

\multirow[c]{3}{=}{\centering\footnotesize\itshape Qwen-7B\\variants}
& Qwen2.5-7B (Base)           & 0.850 & 0.560 \\
&Qwen2.5-7B + CoT           & 0.845 & 0.565 \\
&\textbf{Qwen2.5-7B + DAIS} & \textbf{0.872} & 0.580 \\

\midrule
\multirow[c]{4}{=}{\centering\footnotesize\itshape Reference\\models}
&Qwen2.5-72B                    & 0.898 & 0.68 \\
&GLM-4.7                    & 0.565 & 0.565 \\
&MiniMax-M27                & 0.790 & 0.625\\
&GPT-OSS-120B               & 0.790 & 0.785\\

\bottomrule
\end{tabularx}
\end{table}

\subsection{Full Controlled Results}
\label{app:full_controlled_results}

Table~\ref{tab:main_results_full} reports the full controlled results across four benchmarks and four Qwen backbones.
Scores are averaged over three runs, and the subscripted values denote standard deviations.

\begin{table*}[t]
\centering
\small
\setlength{\tabcolsep}{5pt}
\renewcommand{\arraystretch}{1.03}
\caption{
Accuracy on four benchmarks across four Qwen backbones.
\textsc{IndepQA} removes previous-state conditioning, and \textsc{CR-Data SFT} is an external-data reference for policy tasks.
Bold marks the best score in each column; averages are over backbones.
}\label{tab:main_results_full}
\begin{tabular}{clccccc}
\toprule
\textbf{Benchmark} 
& \textbf{Method} 
& \textbf{Qwen2.5-3B} 
& \textbf{Qwen2.5-7B} 
& \textbf{Qwen3-4B} 
& \textbf{Qwen3-8B} 
& \textbf{Average} \\
\midrule
\multirow{6}{*}{GDPR}
& \textsc{Base}        & 0.756\textsubscript{±0.005} & 0.890\textsubscript{±0.005} & 0.608\textsubscript{±0.005} & 0.768\textsubscript{±0.005} & 0.756\textsubscript{±0.005} \\
& \textsc{Final-SFT}   & 0.788\textsubscript{±0.003} & 0.904\textsubscript{±0.003} & 0.626\textsubscript{±0.003} & 0.770\textsubscript{±0.003} & 0.772\textsubscript{±0.003} \\
& \textsc{CoT-SFT}     & 0.752\textsubscript{±0.006} & 0.896\textsubscript{±0.006} & 0.630\textsubscript{±0.006} & 0.770\textsubscript{±0.006} & 0.762\textsubscript{±0.006} \\
& \textsc{IndepQA}     & 0.686\textsubscript{±0.009} & 0.852\textsubscript{±0.009} & 0.568\textsubscript{±0.009} & 0.750\textsubscript{±0.009} & 0.714\textsubscript{±0.009} \\
& \textsc{CR-Data SFT} & 0.724\textsubscript{±0.005} & 0.912\textsubscript{±0.005} & 0.652\textsubscript{±0.005} & 0.712\textsubscript{±0.005} & 0.750\textsubscript{±0.005} \\
\cmidrule{2-7}
& \textsc{DAIS (Ours)} & \textbf{0.798}\textsubscript{±0.005} & \textbf{0.954}\textsubscript{±0.005} & \textbf{0.656}\textsubscript{±0.005} & \textbf{0.786}\textsubscript{±0.005} & \textbf{0.799}\textsubscript{±0.005} \\
\midrule
\multirow{6}{*}{AIACT}
& \textsc{Base}        & 0.356\textsubscript{±0.005} & 0.408\textsubscript{±0.005} & 0.628\textsubscript{±0.005} & 0.700\textsubscript{±0.005} & 0.523\textsubscript{±0.005} \\
& \textsc{Final-SFT}   & 0.364\textsubscript{±0.003} & 0.404\textsubscript{±0.003} & 0.632\textsubscript{±0.003} & 0.724\textsubscript{±0.003} & 0.531\textsubscript{±0.003} \\
& \textsc{CoT-SFT}     & 0.360\textsubscript{±0.006} & 0.408\textsubscript{±0.006} & 0.636\textsubscript{±0.006} & 0.726\textsubscript{±0.006} & 0.533\textsubscript{±0.006} \\
& \textsc{IndepQA}     & 0.352\textsubscript{±0.009} & 0.418\textsubscript{±0.009} & 0.574\textsubscript{±0.009} & 0.612\textsubscript{±0.009} & 0.489\textsubscript{±0.009} \\
& \textsc{CR-Data SFT} & 0.396\textsubscript{±0.005} & 0.382\textsubscript{±0.005} & 0.654\textsubscript{±0.005} & 0.692\textsubscript{±0.005} & 0.531\textsubscript{±0.005} \\
\cmidrule{2-7}
& \textsc{DAIS (Ours)} & \textbf{0.472}\textsubscript{±0.005} & \textbf{0.486}\textsubscript{±0.005} & \textbf{0.658}\textsubscript{±0.005} & \textbf{0.738}\textsubscript{±0.005} & \textbf{0.589}\textsubscript{±0.005} \\
\midrule
\multirow{5}{*}{MedQA}
& \textsc{Base}        & 0.752\textsubscript{±0.005} & 0.850\textsubscript{±0.005} & 0.602\textsubscript{±0.005} & 0.855\textsubscript{±0.005} & 0.765\textsubscript{±0.005} \\
& \textsc{Final-SFT}   & 0.738\textsubscript{±0.003} & 0.845\textsubscript{±0.003} & 0.732\textsubscript{±0.003} & 0.848\textsubscript{±0.003} & 0.791\textsubscript{±0.003} \\
& \textsc{CoT-SFT}     & 0.752\textsubscript{±0.006} & 0.849\textsubscript{±0.006} & 0.758\textsubscript{±0.006} & \textbf{0.858}\textsubscript{±0.006} & 0.804\textsubscript{±0.006} \\
& \textsc{IndepQA}     & 0.722\textsubscript{±0.009} & 0.818\textsubscript{±0.009} & 0.750\textsubscript{±0.009} & 0.850\textsubscript{±0.009} & 0.785\textsubscript{±0.009} \\
\cmidrule{2-7}
& \textsc{DAIS (Ours)} & \textbf{0.764}\textsubscript{±0.005} & \textbf{0.872}\textsubscript{±0.005} & \textbf{0.769}\textsubscript{±0.005} & 0.856\textsubscript{±0.005} & \textbf{0.815}\textsubscript{±0.005} \\
\midrule
\multirow{5}{*}{FOLIO}
& \textsc{Base}        & 0.470\textsubscript{±0.018} & 0.560\textsubscript{±0.018} & 0.710\textsubscript{±0.018} & 0.835\textsubscript{±0.018} & 0.644\textsubscript{±0.018} \\
& \textsc{Final-SFT}   & 0.475\textsubscript{±0.005} & 0.565\textsubscript{±0.005} & 0.795\textsubscript{±0.005} & 0.795\textsubscript{±0.005} & 0.658\textsubscript{±0.005} \\
& \textsc{CoT-SFT}     & 0.485\textsubscript{±0.013} & 0.570\textsubscript{±0.013} & 0.795\textsubscript{±0.013} & 0.805\textsubscript{±0.013} & 0.664\textsubscript{±0.013} \\
& \textsc{IndepQA}     & 0.473\textsubscript{±0.003} & 0.565\textsubscript{±0.003} & 0.785\textsubscript{±0.003} & 0.795\textsubscript{±0.003} & 0.655\textsubscript{±0.003} \\
\cmidrule{2-7}
& \textsc{DAIS (Ours)} & \textbf{0.500}\textsubscript{±0.005} & \textbf{0.580}\textsubscript{±0.005} & \textbf{0.815}\textsubscript{±0.005} & \textbf{0.840}\textsubscript{±0.005} & \textbf{0.684}\textsubscript{±0.005} \\
\bottomrule
\end{tabular}
\end{table*}

\subsection{Full Data-Efficiency Results}
\label{app:full_data_efficiency}

Tables~\ref{tab:data_efficiency} and~\ref{tab:data_efficiency_medqa} report the full numerical results for the low-resource data-efficiency analysis.
We evaluate Qwen2.5-7B on GDPR and MedQA using different fractions of the original supervised training instances.
The untuned \textsc{Base} model is repeated across fractions as a reference, while the other methods are fine-tuned with the corresponding amount of original data.

On GDPR, \dais{} achieves the best score at every training fraction.
With only 25\% of the training data, \dais{} reaches 0.942, which is already substantially higher than the full-data \textsc{Final-SFT} and \textsc{CoT-SFT} baselines.
On MedQA, the pattern is more conservative: \dais{} ties \textsc{CoT-SFT} among fine-tuned methods at 10\%, remains below the untuned base at that point, and becomes the best method from 25\% onward.
These results suggest that dependency-conditioned intermediate supervision can improve efficiency with respect to original supervised instances, with a stronger effect on the policy-compliance task than on the medical QA task.
They should not be interpreted as equal-budget results in total records, tokens, wall-clock time, or optimization steps.

\begin{table}[!]
\centering
\small
\caption{
Data efficiency on GDPR with Qwen2.5-7B.
\textsc{DAIS} achieves the best result at every training fraction and reaches strong performance with substantially less training data.
}
\label{tab:data_efficiency}
\renewcommand{\arraystretch}{1.05}
\begin{tabular}{lcccc}
\toprule
\textbf{Method} & \textbf{10\%} & \textbf{25\%} & \textbf{50\%} & \textbf{100\%} \\
\midrule
Base      & 0.890 & 0.890 & 0.890 & 0.890 \\
FINAL-SFT       & 0.888 & 0.866 & 0.862 & 0.904 \\
CoT-SFT       & 0.892 & 0.896 & 0.894 & 0.896 \\
DAIS & \textbf{0.896} & \textbf{0.942} & \textbf{0.950} & \textbf{0.954} \\
\bottomrule
\end{tabular}
\end{table}

\begin{table}[!]
\centering
\small
\caption{
Low-resource evaluation on MedQA with Qwen2.5-7B.
Dependency-aware supervision (\textsc{DAIS}) consistently outperforms final-only and unordered intermediate supervision across all training fractions.
}
\renewcommand{\arraystretch}{1.2}
\begin{tabular}{lcccc}
\toprule
\textbf{Method} & \textbf{10\%} & \textbf{25\%} & \textbf{50\%} & \textbf{100\%} \\
\midrule
Base      & 0.850 & 0.850 & 0.850 & 0.850 \\
SFT       & 0.816 & 0.812 & 0.828 & 0.845 \\
CoT       & 0.832 & 0.832 & 0.838 & 0.849 \\
DAIS      & \textbf{0.832} & \textbf{0.852} & \textbf{0.854} & \textbf{0.872} \\
\bottomrule
\end{tabular}
\label{tab:data_efficiency_medqa}
\end{table}

\section{Construction Prompts and Data Examples}
\label{app:construction_prompts_examples}

This section provides the construction prompts and representative data examples used in our supervised target construction.
We include the essential prompt templates rather than implementation-specific file paths.
All examples are shortened for readability; the complete serialized records are provided in the supplementary material.
For readability, the MedQA examples are translated into English, while the constructed training records preserve the original dataset language.

The construction process contains both LLM-based and deterministic components.
Teacher rationales, subtask decompositions, and schema proposals are generated with LLM prompts.
Target serialization, previous-state insertion, and ablation variants are implemented deterministically.
For GDPR and AIACT, we use a predefined compliance-oriented schema because the two datasets share a stable legal-compliance reasoning structure.
For MedQA and FOLIO, we induce dataset-level schemas from generated subtasks and manually audit the resulting categories.

\begin{table*}[t]
\centering
\small
\caption{
Teacher-rationale generation prompt.
The same general instruction is adapted to the answer format of each dataset.
}
\label{tab:teacher_generation_prompt}
\begin{tabularx}{\textwidth}{p{0.22\textwidth}X}
\toprule
Field & Prompt specification \\
\midrule
Model & \texttt{DeepSeek-V4-Flash} for the first candidate rationale. \\
\midrule
Decoding & Temperature $0.7$ for construction-time API calls. \\
\midrule
General prompt &
Generate a concise step-by-step rationale and a final answer from the original task input, optional context or evidence, and answer options when available.
Use only the supplied information.
The rationale should be grounded in the input, decomposable into local reasoning steps, and free of unsupported facts or stylistic filler. \\
\midrule
Policy-compliance format &
Return a rationale explaining the relevant facts, applicable legal or policy conditions, and the final norm judgment.
The final answer must be one of \texttt{permit}, \texttt{prohibit}, or \texttt{unrelated}. \\
\midrule
MedQA format &
Return a concise medical rationale, the final answer text, and the final option letter.
The rationale should focus on diagnostic, clinical, anatomical, physiological, or treatment-selection clues. \\
\midrule
FOLIO format &
Return a concise logic rationale grounded only in the supplied natural-language premises.
The final answer must be one of \texttt{True}, \texttt{False}, or \texttt{Unknown}. \\
\bottomrule
\end{tabularx}
\end{table*}

\begin{table*}[t]
\centering
\small
\caption{
Rationale validation and retry protocol.
This step filters noisy teacher rationales before subtask construction.
}
\label{tab:validation_retry_prompt}
\begin{tabularx}{\textwidth}{p{0.22\textwidth}X}
\toprule
Field & Prompt specification \\
\midrule
Validation model & \texttt{DeepSeek-V4-Pro}. \\
\midrule
Retry model & \texttt{DeepSeek-V4-Pro}. \\
\midrule
Validation instruction &
Check whether the candidate rationale should be retained for training-data construction.
A retained candidate must satisfy two conditions: its predicted final answer matches the gold answer, and its non-trivial reasoning steps are grounded in the input, context, answer options, or preceding reasoning steps. \\
\midrule
Rejection criteria &
Reject the candidate if it contains unsupported reasoning, explicit contradictions, reasoning that supports an alternative answer, or a final conclusion inconsistent with the stated answer. \\
\midrule
Retry rule &
If the first candidate fails validation, regenerate the rationale with \texttt{DeepSeek-V4-Pro} for up to two additional rounds.
We retain the first valid candidate.
If no candidate passes, the instance is removed from the trace-construction set. \\
\bottomrule
\end{tabularx}
\end{table*}

\begin{table*}[t]
\centering
\small
\caption{
Subtask generation and DAIS QA construction instruction.
For GDPR and AIACT, the schema is predefined; for MedQA and FOLIO, subtasks are generated from filtered teacher rationales.
}
\label{tab:subtask_generation_prompt}
\begin{tabularx}{\textwidth}{p{0.22\textwidth}X}
\toprule
Field & Prompt specification \\
\midrule
Model & \texttt{DeepSeek-V4-Pro}. \\
\midrule
Decoding & Temperature $0.7$ for construction-time API calls. \\
\midrule
General prompt &
Decompose the retained rationale into a compact ordered sequence of two to five local QA subtasks.
Each subtask must contain a local question, the information needed to answer it, concise reasoning, an intermediate answer, and a reasoning-function label.
Subtasks must be decision-relevant, non-redundant, and grounded in the original input and teacher rationale. \\
\midrule
MedQA-specific instruction &
Create local QA subtasks in the same language as the original question.
Each subtask must be derived from the original question, options, gold answer, and rationale.
Do not introduce external medical facts beyond the original question and rationale. \\
\midrule
FOLIO-specific instruction &
Create English QA subtasks for logic entailment.
Each subtask must be supported by the premises and rationale.
The sequence should typically include evidence retrieval, local deduction, and final veracity classification. \\
\midrule
GDPR / AIACT instruction &
Use the predefined compliance schema.
Generate subtask-level QA states for legal element extraction, regulation or article mapping, and compliance reasoning. \\
\midrule
DAIS construction rule &
For the first subtask, construct the local question from the original input and context.
For later subtasks, include a \texttt{Previous states:} block that serializes earlier subtask states as context for the current decision.
The final-answer record uses only the original task input and optional context, without constructed intermediate states. \\
\bottomrule
\end{tabularx}
\end{table*}

\begin{table*}[t]
\centering
\small
\caption{
Schema induction instruction for MedQA and FOLIO.
GDPR and AIACT use a predefined compliance schema and do not require clustering.
}
\label{tab:schema_prompt}
\begin{tabularx}{\textwidth}{p{0.22\textwidth}X}
\toprule
Field & Prompt specification \\
\midrule
Models & \texttt{DeepSeek-V4-Pro}, followed by manual audit. \\
\midrule
Decoding & Temperature $0.7$ for construction-time API calls. \\
\midrule
Prompt &
Given noisy free-form subtask labels, label counts, and representative examples, cluster the labels into a compact set of reusable reasoning categories.
The categories should describe reasoning functions rather than surface wording, entities, answer strings, or dataset-specific content. \\
\midrule
Output requirement &
Return a compact schema in which each raw label is assigned to exactly one canonical category.
Each category should have a concise name and a short description. \\
\midrule
Manual audit &
After LLM-assisted clustering, we merge overlapping types, revise overly broad or ambiguous categories, correct inconsistent assignments, and relabel the generated subtasks according to the finalized schema. \\
\bottomrule
\end{tabularx}
\end{table*}

\begin{table*}[t]
\centering
\small
\caption{
Evaluation protocol and answer extraction.
No teacher rationale, gold intermediate state, or external decomposition module is provided at evaluation time.
}
\label{tab:evaluation_prompt}
\begin{tabularx}{\textwidth}{p{0.22\textwidth}X}
\toprule
Field & Prompt specification \\
\midrule
General evaluation prompt &
Answer the original task input using the same task format across methods.
The main evaluation asks the model to produce the final answer directly. \\
\midrule
Policy-compliance evaluation &
Judge whether the case is \texttt{permit}, \texttt{prohibit}, or \texttt{unrelated}.
The predicted norm type is extracted from the model output, normalized by the fixed answer matcher when needed, and compared with the gold label. \\
\midrule
MedQA evaluation &
Answer the medical multiple-choice question and provide the final answer.
When an option letter is present, the normalized predicted option is matched against the gold option. \\
\midrule
FOLIO evaluation &
Determine whether the conclusion is \texttt{True}, \texttt{False}, or \texttt{Unknown}.
The normalized final label is matched against the gold label. \\
\midrule
Decoding &
Most local evaluation uses deterministic decoding with temperature $0$.
Long-form reasoning models use stochastic decoding as described in Appendix~\ref{app:decoding_evaluation}. \\
\bottomrule
\end{tabularx}
\end{table*}

All supervised variants are represented as instruction--input--output records.
The main difference between \textsc{IndepQA} and \dais{} is the treatment of previous intermediate states.
Both formats expose local QA states, but only \dais{} inserts earlier subtask outputs into the input of later intermediate subtasks.
The final-answer record remains the original final-answer task.
The main markers include \texttt{[Stage i | label]}, \texttt{Current subtask question:}, \texttt{Current subtask input:}, \texttt{Previous states:}, \texttt{Answer:}, and \texttt{Final option:}.
Table~\ref{tab:target_format_comparison} summarizes the resulting target-format difference.
\dais{} makes support relations visible through input serialization rather than through explicit dependency-edge prediction.
Therefore, the dependency structure should be interpreted as a target-construction device for SFT, not as a separately supervised graph parser.

\begin{table*}[t]
\centering
\small
\caption{
Target-format comparison.
\dais{} differs from \textsc{IndepQA} by inserting previous subtask outputs into the input of later intermediate subtasks.
The final-answer record remains the original task.
}
\label{tab:target_format_comparison}
\begin{tabularx}{\textwidth}{p{0.17\textwidth}p{0.38\textwidth}X}
\toprule
Target format & Input context & Target output \\
\midrule
\textsc{Final-SFT} &
Original task input only. &
Final answer only. \\
\midrule
\textsc{CoT-SFT} &
Original task input only. &
Flat teacher rationale followed by the final answer. \\
\midrule
\textsc{IndepQA} &
Original task input and the current local subtask question.
No \texttt{Previous states:} block is included. &
Current subtask answer for intermediate records; final answer for the final record. \\
\midrule
\dais{} intermediate record &
Original task input, current local subtask question, and a \texttt{Previous states:} block for later subtasks.
The block serializes earlier subtask questions, inputs, reasoning, and answers. &
Current subtask answer. \\
\midrule
\dais{} final-answer record &
Original task input only. &
Final answer. \\
\bottomrule
\end{tabularx}
\end{table*}

The following examples instantiate this format for GDPR, MedQA, and FOLIO.
They cover the three main reasoning domains in our experiments.
AIACT follows the same compliance-oriented target structure as GDPR and is summarized in Table~\ref{tab:subtask_examples_all}.

\begin{table*}[t]
\centering
\small
\caption{
Representative GDPR target example.
In \dais{}, later compliance subtasks receive previous subtask outputs as part of their input; the final-answer record uses only the original case.
}
\label{tab:gdpr_target_example}
\begin{tabularx}{\textwidth}{p{0.15\textwidth}p{0.42\textwidth}X}
\toprule
Format / record & Input contains & Target output \\
\midrule
Original QA &
Case: Meta Platforms Ireland Limited stored user passwords internally without encryption and did not report or document a related data breach.
Question: judge whether the case is \texttt{permit}, \texttt{prohibit}, or \texttt{unrelated}. &
Gold answer: \texttt{prohibit}. \\
\midrule
\textsc{Final-SFT} &
Original case only. &
\texttt{prohibit}. \\
\midrule
\textsc{CoT-SFT} &
Original case only. &
A flat rationale explaining that unencrypted password storage violates security obligations and that failure to report or document the breach implicates breach-notification duties.
Final answer: \texttt{prohibit}. \\
\midrule
\dais{} Stage 1: Legal Element Extraction &
Original case and current subtask question:
extract legal actors, roles, information type, consent form, and processing purpose. &
Controller = Meta Platforms Ireland Limited;
data subjects = users;
information type = passwords;
context = internal systems;
purpose = data storage. \\
\midrule
\dais{} Stage 2: Legal Regulation Mapping &
Original case and current subtask question.
\texttt{Previous states:} Stage 1 output with extracted legal elements. &
Violated articles: Article 32, Article 33, and Article 34. \\
\midrule
\dais{} Stage 3: Compliance Reasoning &
Original case and current subtask question.
\texttt{Previous states:} Stage 1 extracted elements and Stage 2 violated articles. &
Norm type: \texttt{prohibit};
violated articles: Article 32, Article 33, and Article 34. \\
\midrule
\dais{} Final-answer record &
Original case only. &
\texttt{prohibit}. \\
\bottomrule
\end{tabularx}
\end{table*}

\begin{table*}[t]
\centering
\small
\caption{
Representative MedQA target example.
The second \dais{} subtask conditions on the diagnosis output by the first subtask; the final-answer record uses only the original question and options.
}
\label{tab:medqa_target_example}
\begin{tabularx}{\textwidth}{p{0.15\textwidth}p{0.42\textwidth}X}
\toprule
Format / record & Input contains & Target output \\
\midrule
Original QA &
Question: A 50-year-old man develops sudden severe pain, redness, and swelling in the first metatarsophalangeal joint at night after eating seafood.
Options: A. Benzbromarone; B. Allopurinol; C. Antibiotics; D. NSAIDs; E. Methotrexate. &
Gold answer: NSAIDs; final option: D. \\
\midrule
\textsc{Final-SFT} &
Original question and options. &
Answer: NSAIDs; final option: D. \\
\midrule
\textsc{CoT-SFT} &
Original question and options. &
Flat rationale: acute first metatarsophalangeal joint pain after a high-purine meal suggests acute gouty arthritis; acute attacks are treated with NSAIDs, while allopurinol and benzbromarone are for chronic urate lowering.
Final answer: NSAIDs; option D. \\
\midrule
\dais{} Stage 1: Clinical Decision Making &
Original question, options, and current subtask question:
based on the clinical presentation, what is the most likely diagnosis? &
Acute gouty arthritis. \\
\midrule
\dais{} Stage 2: Clinical Decision Making &
Original question, options, and current subtask question:
which drug class should be selected for an acute gout attack?
\texttt{Previous states:} Stage 1 output = acute gouty arthritis. &
NSAIDs. \\
\midrule
\dais{} Final-answer record &
Original question and options only. &
Answer: NSAIDs; final option: D. \\
\bottomrule
\end{tabularx}
\end{table*}

\begin{table*}[t]
\centering
\small
\caption{
Representative FOLIO target example.
Later logical subtasks receive earlier derived cases and local conclusions as input; the final-answer record uses only the original premises and conclusion.
}
\label{tab:folio_target_example}
\begin{tabularx}{\textwidth}{p{0.15\textwidth}p{0.42\textwidth}X}
\toprule
Format / record & Input contains & Target output \\
\midrule
Original QA &
Premises about Rina, coffee drinking, caffeine dependence, joking about caffeine addiction, and being unaware that caffeine is a drug.
Conclusion: Rina jokes about being addicted to caffeine or is unaware that caffeine is a drug. &
Gold answer: \texttt{True}. \\
\midrule
\textsc{Final-SFT} &
Original premises and conclusion. &
\texttt{True}. \\
\midrule
\textsc{CoT-SFT} &
Original premises and conclusion. &
Flat rationale: the premises yield two cases for Rina; in the first she is unaware, and in the second she must joke about being addicted to caffeine. Since the conclusion holds in both cases, the answer is true. \\
\midrule
\dais{} Stage 1: Evidence Retrieval &
Original premises and current subtask question:
what are the two possible cases for Rina? &
Case 1: Rina is a student, unaware, and dependent on caffeine.
Case 2: Rina is not a student, not unaware, and not dependent on caffeine. \\
\midrule
\dais{} Stage 2: Local Deduction &
Original premises and current subtask question:
does the conclusion hold in each case?
\texttt{Previous states:} Stage 1 output with the two cases. &
Yes. The conclusion holds in both cases. \\
\midrule
\dais{} Stage 3: Veracity Classification &
Original premises and current subtask question:
does the conclusion necessarily follow?
\texttt{Previous states:} Stage 1 case split and Stage 2 local deduction output. &
\texttt{True}. \\
\midrule
\dais{} Final-answer record &
Original premises and conclusion only. &
\texttt{True}. \\
\bottomrule
\end{tabularx}
\end{table*}

Table~\ref{tab:subtask_examples_all} reports representative subtask schemas for all four datasets.
Unlike a pure schema table, it also shows the typical previous-state outputs that are inserted into the input of later \dais{} subtask records.

\begin{table*}[t]
\centering
\small
\caption{
Representative subtask labels, local questions, and dependency contexts.
The final column shows which previous outputs are typically inserted into the input of the current \dais{} subtask record.
}
\label{tab:subtask_examples_all}
\begin{tabularx}{\textwidth}{p{0.10\textwidth}p{0.22\textwidth}p{0.34\textwidth}X}
\toprule
Dataset & Subtask label & Example local question & Typical previous-state input in \dais{} \\
\midrule
GDPR &
Legal Element Extraction &
Extract the legal actors, roles, information type, consent form, and processing purpose. &
None; first-stage extraction from the case. \\
\cmidrule(lr){2-4}
&
Legal Regulation Mapping &
Map the case or extracted legal elements to the relevant GDPR articles. &
Output of Legal Element Extraction. \\
\cmidrule(lr){2-4}
&
Compliance Reasoning &
Determine whether the case is permitted, prohibited, or unrelated. &
Outputs of Legal Element Extraction and Legal Regulation Mapping. \\
\midrule
AIACT &
Legal Element Extraction &
Extract the provider, user, affected subjects, AI-system function, information type, and deployment purpose. &
None; first-stage extraction from the case. \\
\cmidrule(lr){2-4}
&
Legal Regulation Mapping &
Map the extracted AI-system elements to applicable AI Act or policy-compliance provisions. &
Output of Legal Element Extraction. \\
\cmidrule(lr){2-4}
&
Compliance Reasoning &
Determine whether the AI-system deployment is permitted, prohibited, or unrelated. &
Outputs of Legal Element Extraction and Legal Regulation Mapping. \\
\midrule
MedQA &
Clinical Evidence Extraction &
What clinical findings, symptoms, laboratory results, or temporal clues are relevant? &
None or earlier clinical context only. \\
\cmidrule(lr){2-4}
&
Clinical Concept Mapping &
Which disease, mechanism, drug class, or medical concept is indicated by the evidence? &
Often uses extracted clinical evidence. \\
\cmidrule(lr){2-4}
&
Differential \& Constraint Reasoning &
Which option is compatible with the clinical constraints, and which alternatives should be eliminated? &
Often uses evidence and mapped clinical concepts. \\
\cmidrule(lr){2-4}
&
Mechanistic / Causal Inference &
What pathophysiological or causal mechanism explains the phenomenon? &
Often uses extracted evidence and concept-mapping outputs. \\
\cmidrule(lr){2-4}
&
Clinical Decision Making &
What diagnosis, treatment, examination, or management step should be selected? &
Often uses previous diagnosis, mechanism, or option-elimination outputs. \\
\midrule
FOLIO &
Evidence Retrieval &
Which premises or cases are relevant to the conclusion? &
None; first-stage retrieval from premises. \\
\cmidrule(lr){2-4}
&
Local Deduction &
What local consequence follows from the selected premises or case split? &
Output of Evidence Retrieval. \\
\cmidrule(lr){2-4}
&
Multi-hop Deduction &
How do multiple local conclusions combine to support or refute the target conclusion? &
Outputs of earlier retrieval and local deduction states. \\
\cmidrule(lr){2-4}
&
Veracity Classification &
Does the conclusion follow as \texttt{True}, \texttt{False}, or \texttt{Unknown}? &
All relevant previous intermediate outputs. \\
\bottomrule
\end{tabularx}
\end{table*}

Finally, several construction components are implemented procedurally rather than as separate prompts.
Rationale-span alignment is represented through the subtask input and reasoning fields rather than character-offset annotations.
Intermediate answers are emitted as subtask-answer fields during decomposition or are derived from structured compliance fields for GDPR and AIACT.
Dependency conditioning is implemented by serializing previous states into later subtask inputs, not by training the model to predict explicit dependency-edge tokens.
The final-answer record remains the original task input paired with the gold final answer.
This design keeps the supervised target textual while making the support relation from earlier outputs to later local decisions visible during training.

\end{document}